\documentclass[10pt,twocolumn,letterpaper]{article}

\usepackage{cvpr} 

\usepackage{cite}
\usepackage{amsmath,amssymb,amsfonts}
\usepackage{algorithmic}
\usepackage{graphicx}
\usepackage{textcomp}
\usepackage{xcolor}
\usepackage{url}
\usepackage{verbatim}
\usepackage{multirow}
\usepackage{comment}
\usepackage[title]{appendix}

\newcommand{\TopOne}{78.4}

\newcommand{\TopThree}{94.4}
\newcommand{\TopOneChance}{4}
\newcommand{\TopThreeChance}{12}
\newcommand{\SampleRate}{125}
\newcommand{\SegLen}{100}
\newcommand{\SegOffset}{10}
\newcommand{\NumRawChan}{16}
\newcommand{\NumChan}{14}
\newcommand{\FilterSize}{3}
\newcommand{\NumFilters}{256}
\newcommand{\EncodingSize}{32}
\newcommand{\ProjHeadSize}{64}
\newcommand{\ModelSize}{655,136}
\newcommand{\KNearest}{25}
\newcommand{\BatchSize}{8}
\newcommand{\NumConcepts}{103}
\newcommand{\NumWikiPages}{150}
\newcommand{\WikiPageGroupSize}{10}
\newcommand{\NumTestConcepts}{25}
\newcommand{\NumTrainingConcepts}{78}
\newcommand{\NumEpochs}{8}
\newcommand{\StepsPerEpoch}{500}
\newcommand{\NumTrials}{25}

\begin{document}
\title{Neural Memory Decoding with\\EEG Data and Representation Learning}

\author{Glenn Bruns\qquad Michael Haidar \qquad Federico Rubino\\
California State University, Monterey Bay\\
Seaside, CA 93955
}

\maketitle

\begin{abstract}
We describe a method for the neural decoding of memory from EEG data.
Using this method, a concept being recalled can be 
identified from an EEG trace with an average top-1 accuracy of about {\TopOne}\%
(chance {\TopOneChance}\%).  
The method employs deep representation learning with supervised contrastive
loss to map an EEG recording of brain activity to a low-dimensional space.
Because representation learning is used, concepts can be identified even if they
do not appear in the training data set.  However, reference EEG data must
exist for each such concept.
We also show an application of the method to the problem of information retrieval.
In neural information retrieval, EEG data is captured while a user recalls the
contents of a document, and a list of links to predicted documents is produced.
\end{abstract}

\section{Introduction} 
\label{sec:intro}

\begin{figure}[ht]
\begin{center}
\includegraphics[scale=.33]{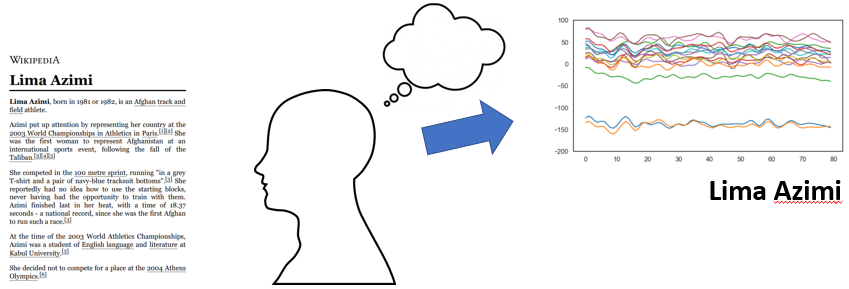}

\vspace{12pt}

\includegraphics[scale=.33]{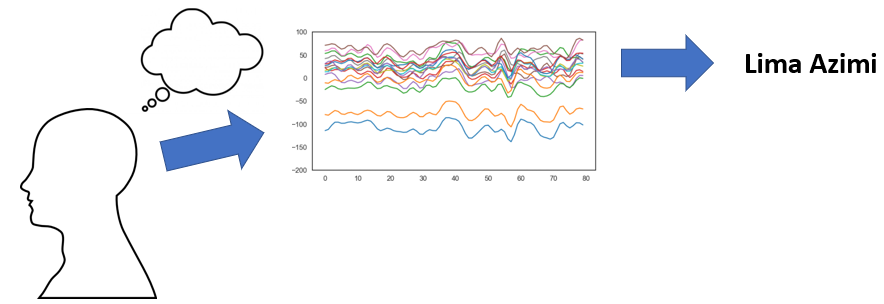}
\end{center}
\caption{A problem of neural memory decoding.  {\bf Top}: An individual reads about a concept.  Brain activity is recorded and labeled with the concept name.  {\bf Bottom}: Later, the individual recalls the concept.  Brain activity is collected and used to identify the concept.}
\label{fig:concept}
\end{figure}

Neural decoding is the reconstruction of stimuli or mental state from a record of 
electrical activity in the brain.  An example is the reconstruction of an 
individual's emotional state (happy, sad, or neutral) from electroencephalogram (EEG) data.
In the case of neural {\em memory} decoding, the problem is to identify a concept
that is being recalled from a record of electrical activity in the brain. 
The problem is illustrated in Fig.\ref{fig:concept}. 
Here we show it is possible, with high accuracy, to identify a concept
being recalled using EEG data.  It is assumed that a previously-recorded 
reference EEG recording is available for every concept to be recalled.

The neural memory decoding problem is important because of the light a solution
can shed on the function of human memory.  For example, consider the question
of which frequency bands (alpha, gamma, etc.) are most associated with memory
function.  In a direct approach to understanding this relationship, one can
examine brain data collected while a subject performs memory-related tasks.
With a system for neural memory decoding, an alternative method can be
used in which the accuracy of memory decoding is compared across 
multiple data sets, each collected from a different set of frequency bands.
Similarly, to help understand which regions of the brain are associated with 
memory function, one can compare the accuracy of memory decoding across 
data sets containing data from different EEG nodes.

The neural memory decoding problem is also important because a solution
to the problem would enable useful applications.
For example, a neural memory decoding system
could be used for {\em neural document retrieval}, in which a user
retrieves a previously-read document by simply thinking about it.  

Solving the neural memory decoding problem is challenging.
Brain structure changes over time.  The brain activity associated with 
recalling something today will be different than recalling it tomorrow.  
Also, memory is a complex mental activity involving multiple cognitive 
functions and regions of the brain.  Recalling a concept does not involve a simple, localized pattern of brain activity.  

Another challenge in solving the problem of neural memory decoding is 
in the experimental design and data organization.  For each of many concepts, brain data must be recorded at multiple time points.  
Predicting a concept being recalled involves relating data recorded
at the time of recall to data recorded at an earlier time. Additionally, while EEG recording devices are preferred for cost-effective experimentation, they suffer from noise and limited spatial resolution in comparison to more expensive alternatives like Functional Magnetic Resonance Imaging (fMRI).

In this paper, we demonstrate the feasibility of solving the neural 
memory decoding problem.  
Our data set consists of EEG traces from a single subject for {\NumConcepts} concepts.
For each concept, EEG data was recorded while the subject recalled the concept
at three points in time:  immediately after the concept was learned,  
one day after the concept was learned, and 3 days after the concept
was learned.

Our system correctly identifies the concept the subject recalls on
day one with about \TopOne{}\% accuracy, on average, when trained on 
EEG traces for {\NumTrainingConcepts} concepts and tested on EEG traces 
for {\NumTestConcepts} concepts.  An average accuracy of only {\TopOneChance}\% would be 
achieved by chance.  Our system achieves a top-3 accuracy (the correct
concept is among the top three predicted concepts) of about \TopThree{}\% 
(versus {\TopThreeChance}\% by chance).

In operation, our system takes as input a \NumRawChan{}-channel EEG trace 
recorded on day 1 and produces as output the concept name.  The system
is a deep neural net.  It predicts the concept of a trace by segmenting 
the input EEG trace, mapping each EEG segment to a point in a compact 
representation space, predicting the concept of each point, and then combining 
these predictions to obtain a concept for the EEG trace as a whole.

The use of representation learning allows the concept of an EEG trace to
be predicted even if no EEG trace with that concept was seen during the training
of the system.  This property is important in using the system for applications
such as neural document retrieval.  In this application, no new training is
needed to classify a new concept.

In experiments that test system performance on data sets containing only
a subset of the EEG nodes or frequency bands, we found that the left frontal
and central nodes provided the highest trace classification accuracy of all
individual nodes, and that the theta and gamma frequency bands provided the
highest trace classification accuracy among all individual frequency bands.

Summarizing our main contributions, we have defined a problem of neural memory decoding
with EEG data.  We have developed a system that achieves top-1 concept prediction
accuracy of about \TopOne{}\% when used on EEG data recorded when recalling
a concept one day after it was learned.  We have used our system
to measure the predictive power of individual EEG nodes and frequency bands.

\section{Related work}
\label{sec:related}

Advancements in neuroimaging technology and computing have accelerated work in predicting the stimuli responsible for neural activity. FMRIs have emerged as the preferred method for investigating semantic decoding, a process that seeks to discern an individual's thoughts,  but not necessarily memories. FMRI surpasses EEG in spatial resolution and functional detail, with neural patterns showing greater spatial distinctiveness compared to temporal characteristics. Additionally, EEG has a lower signal-to-noise ratio (SNR). However, 
machine learning and deep learning offer significant improvement in feature extraction, enabling scientists to effectively enhance the SNR and overcome the inherent low SNR associated with EEG data. \cite{Rybar2022}. Higher SNR is desired for lower levels of background noise. EEGs have also demonstrated comparable effectiveness in neural decoding, as shown by their ability to identify the subject of photos pertaining 
to either land mammals or tools \cite{murphy2011, simanova2010}. These two 
studies provide evidence of category-specific oscillatory patterns during semantic 
processes. However, unlike in this work, neither used deep learning nor did 
they explore multi-day semantic memory.

Several 
studies demonstrate machine learning's ability to decode cognitive functions 
from neural activity recorded via EEG. Emotion recognition, which seeks to 
identify emotional states \cite{ding1997, bird2019, yang2018, rayatdoost2018, li2018}, 
has applications ranging from clinical research \cite{mayor2021} to targeted 
product development \cite{liu2013,eugster2016natural}. Other studies involving 
neural decoding via EEG and machine learning involve motor tasks and motor 
imagery \cite{cruz2014,kumar2016, schirrmeister2017, tayeb2019, li2019decoding, halme2018, sun2021}, seizure detection \cite{fergus2015, murugavel2016, mardini2020, savadkoohi2020},
mental workload \cite{dehais2020, zhou2021}, and even visual stimulus decoding
\cite{shen2019deep, jiao2019decoding}. 

Semantic neural decoding attempts to interpret meaningful information from neural activity associated with language, semantic associations, and meaning-related stimuli.
The underlying cognitive mechanisms involved in understanding and representing meaning are of particular interest. For example, neural activity patterns exhibit similarities when contemplating pictures and words belonging to the same category, providing evidence of unique neural patterns associated with semantic concepts combined with visual stimuli \cite{kumar2017}. Semantic concepts, such as faces, locations, 
and objects also have distinct neural patterns that persist between a 
study and recall phase \cite{polyn2006}. This lends support to reinstatement 
theory, a memory model in which recall occurs through reactivation of 
cortical patterns specific to a memory \cite{JOHNSON2009},  and implies 
that memories can be discriminated categorically. Distinct neural signatures have been demonstrated for abstract concepts such as \emph{multiplication} and \emph{consciousness} \cite{vargas2020}. Using fMRI, participants' brain activity was recorded during contemplation of 28 abstract concepts and subsequently during the recall of previously studied properties. Discrimination of neural signatures was observed across abstract concepts.

Less common is work on the neural decoding of memory. A theoretical 
division of memory separates memory into two sub-types. Episodic memory 
consists of memory involving events or episodes in a person's life, 
whereas semantic memory consists of general knowledge or facts \cite{tulving1972}.
The creation and retrieval of semantic memories is dependent 
on language processing, working memory, and other phonological 
processes \cite{kutas2000}. 

Some previous work has decoded EEG data
of episodic memory.  In \cite{Bramao2018}, a classifier was developed to predict the category of a photograph based on EEG data recorded during the
encoding of an episodic memory involving a photograph and a paired
word.  Each photograph belonged to one of three categories:
landscape, face, or object.  Different authors, using the same data set 
as in \cite{Bramao2018}, achieved a classification
accuracy of 0.74 on the same problem using a Morlet transform network \cite{Keding2021}.

FMRI data and multi-voxel pattern analysis (MVPA) have also been used to classify episodic memories. In one study faces were classified as new (unseen) and old (seen) faces \cite{rissman2010detecting}. Researchers collected data while participants identified whether pictures of faces had been previously studied and successfully decoded explicitly identified faces as new or old, showing individual episodic memories are detectable using neural decoding of memory states. The study was later expanded to include 180 scenes of participants' daily life. FMRI data was again collected, this time participants were asked to categorize a captured video image into six different levels of familiarity from ``confidently not their experience" to ``perfect recollection of experience". The researchers used MVPA to classify between the six recollection statuses with .35 accuracy \cite{rissman2016decoding}.
% Consider including: "Bramao and Johnson’s (2018) findings suggest that the neural correlates of recollection comprise both content-independent and content-specific activity. The overlap between the content-specific activity and the activity engaged when the same content was originally encoded lends strong support to the reinstatement hypothesis."

The work described in this paper differs in three main ways from the 
aforementioned studies.
The first difference is the time at which the training data was collected. 
Our data was collected across multiple days following a study period on 
the first day. The spread of recordings across different days increases the variability of the EEG data, resulting in a more challenging classification task. 
Multi-day data tests the ability of convolutional neural networks (CNNs) to decode 
neural signals used to access (concept-specific) content stored in 
memory while representing a more realistic application of brain-computer interface (BCI) information 
retrieval. Second, both our training and testing data were of cued semantic 
recall (information was presented to cue a memory), there was no verbal response. Whereas \cite{Bramao2018} and \cite{Keding2021} used pairs-associates encoding data recorded during the study phase as 
training data and tested on image-cued verbal recollection data. 
While there is evidence that encoding and recollection processes of memory 
co-vary \cite{Gordon2013}, particularly in hippocampal activation and cortical reinstatement \cite{Johnson2007}, studies able to observe such phenomena historically employ 
more sophisticated equipment such as fMRIs \cite{Johnson2007, Gordon.et.al.2013, Johnson.et.al.2009, Ritchey.et.al.2012}. Third, our system uses CNNs, which are distinct from MVPA. CNNs are a class of deep learning models designed to learn hierarchical representations of data through convolutional layers. 
MVPA is a computational method primarily used to analyze patterns of neural activity across voxels, three-dimensional representations of a specific location within the brain, typically from an fMRI. A voxel contains information on intensity level or neural activity measured at a particular location. 

To our knowledge, no previous study has achieved neural decoding of semantic memory using EEG and deep learning. Beyond semantic memory 
decoding, deep learning has increasingly been used for other neural decoding tasks. Deep learning with a transformer architecture has been used to classify five stages of sleep through EEG representation learning \cite{masked-auto-encoder}. The authors use self-supervised learning with loss measured between raw EEG input and reconstructed signals from masked features (reconstruction loss), while we use supervised contrastive loss.

Multi-modal models have been used in sentiment analysis and relation detection for natural language processing \cite{Hollestein2021}. 
The authors decoded EEG representations using both EEG and eye tracking data. Other recent work in neural decoding and classification using deep 
learning focuses on working memory \cite{zygierewicz2022, HAN2020324}, 
sleep state levels \cite{mohsenvand2020contrastive, banville2021}, and 
abnormality detection \cite{banville2021, mohsenvand2020contrastive}. A few studies have also used multi-input models to work across multiple paradigms such as imaginary movement, visually evoked responses, movement-related cortical potential, and error evoked responses \cite{schirrmeister2017, lawhern2018eegnet}.

\section{Background}

\label{sec:background}
Machine learning has increasingly been used to analyze physiological signals such as heart rate, blood pressure, electrocardiography, and EEG \cite{Acharya2019, Lee2017, schirrmeister2017}. BCIs harness mental activity, often measured via EEG, to directly control a range of devices from prosthetics to computers \cite{Gerven2009}. This mental activity originates from cognitive functions and creates measurable voltage fluctuations across the brain, the result of billions of neuronal events. 

EEG measures the frequency at which neuron potentiation rises and falls - referred to as neural oscillations or brain waves - during brain activity. These brain waves give insight into the communication between neurons and reflect cognitive, and sensory, mental processes \cite{WARD2003}. EEG devices read these measurements via electrodes placed across the scalp. Each electrode (channel) measures the frequency of brain activity at its location. A standard analysis of brain activity uses frequency bands (rates of oscillations) - delta (1–3 Hz), theta (4–7 Hz), alpha (8–12 Hz), beta (13–30 Hz), gamma (30–100 Hz) - and their observed locations \cite{Siegal2012}. Types of brain activities observable with EEG include processing of information that results in alpha-band oscillations \cite{KLIMESCH2012606}. Other cognitive tasks, such as generation and rehearsal of speech from verbal working memory, produce theta frequencies in both ventral-frontal and left parieto-temporal areas of the brain \cite{HARMONY199925}. Similar areas are also activated during cued recall of explicit memory (memories of facts) \cite{ALLAN1997}. Despite an established understanding of frequencies and topology, the high intra-subject variability of brain activity has plagued EEG studies \cite{MELTZER2007}.

Other methods of investigating brain activity include functional magnetic resonance imaging (fMRI), positron emission tomography (PET), magnetoencephalography (MEG), and functional near-infrared spectroscopy (fNIRS). 
Currently, only fNIRS and EEG provide the temporal resolution, cost-effectiveness, and portability desired in a BCI system. For our purposes, EEG 
was selected for its cost and wealth of relevant literature.  

\section{Data collection and preprocessing}
\label{sec:data}

\subsection{Data Collection} 

Raw data was collected with OpenBCI's Ultracortex Mark IV Headset,  
Cyton board, Daisy module, and GUI software (version v5.0.0). 
EEG data was sampled at {\SampleRate} Hz using 16 channels 
organized in the 10-20 system with a referential montage.  
The earlobe was used as an online reference signal.

{\NumWikiPages} Wikipedia \cite{wiki:main} pages were selected 
% by one of the paper's authors 
and arranged into groups of size {\WikiPageGroupSize}.  
The topics were chosen to be neither 
too obscure nor too familiar. Many concern 
people (e.g., Joan Didion, Alan Greenspan), 
places (e.g., Copenhagen, Lake Victoria), and 
ideas (e.g., gravity assist, optical illusion).
In what follows, Wikipedia page topics are referred to as
{\em concepts}.

Each recording session involved one group of concepts. 
The day 0 recording session for each group was structured as follows. First, the participant (an undergraduate student) studied the Wikipedia pages for the concepts in the group.
Each page was studied for five minutes.
The participant was then cued and recorded while mentally recalling the main points of the page. Each recording was 75 seconds long.
The day 1 recording session for a group took place the following day and had 
the same structure as the day 0 session, excluding the 5 minute study period.
The day 3 recording session took place two days after the day 
1 recording session and had the same structure as the day 1 session.
The participant was not involved in the selection of the concepts.

The recordings for 47 pages were dropped because of data quality issues,
leaving recordings for {\NumConcepts} pages, which are listed in
Appendix \ref{sec:appendix}.

\subsection{Preprocessing}

We refer to the EEG data collected for one concept at a single
sitting as a {\em trace}.  Each raw trace was preprocessed using 
the following sequence of operations:

\textbf{Clipping}. Sample values outside the range defined by
the 0.005 and 0.995 quantile values (computed for the trace 
as a whole) are replaced by the range limit values.  This process
is also known as ``clamping''.

\textbf{Occipital Noise Removal}.
For each channel, at each time step, the sample value of the channel 
is reduced by the sample value of the occipital channel, ipsilaterally. The left occipital channel serves as the new reference for the 
left-hand channels, and similarly for the right-hand channels. 
The occipital channels are then removed from the trace, leaving 14 channels.

While other re-referencing methods appear in the literature, 
there is no definitive method for increasing memory signal, a process of removing unwanted noise from EEG data containing memory information.
The occipital electrodes (O1, O2) are closest to Brodmann's 
area 17/18 corresponding to the visual cortex and are thought to 
carry more noise from the visual network \cite{Lei2017, Hillyard781}. 
Re-referencing with the occipital nodes allows for stronger signals 
recorded at regions of the brain associated with memory (see Section \ref{subsec:node-freq-importance}).

\textbf{Frequency filtering}.
A FIR bandpass filter is applied to retain data only in 
1-100 Hz range, which includes the delta, 
theta, alpha, beta, and gamma frequency bands.

\textbf{Normalization}.
Each channel in the trace is independently normalized using 
z-score normalization, so that the samples of the channel 
are zero-centered and have unit standard deviation.

\textbf{Trimming}.
Finally, the first and last 4 seconds of the trace are removed.  
This step removes edge artifacts produced by the FIR filter 
and removes data recorded while the subject relaxes and begins 
focusing on a concept.

\section{Method}
\label{sec:method}

Our system is a multi-class classifier that takes 
an EEG trace as input and produces a probability distribution over concept 
classes as output.  Key features of the system are the use of representation
learning, the treatment of data collected at different points in time, and
the ensemble method used to classify EEG traces using predictions of the
segments in the trace.

\subsection{Overview}
\label{subsec:overview}

\begin{figure}[ht]
\begin{center}
\includegraphics[scale=.42]{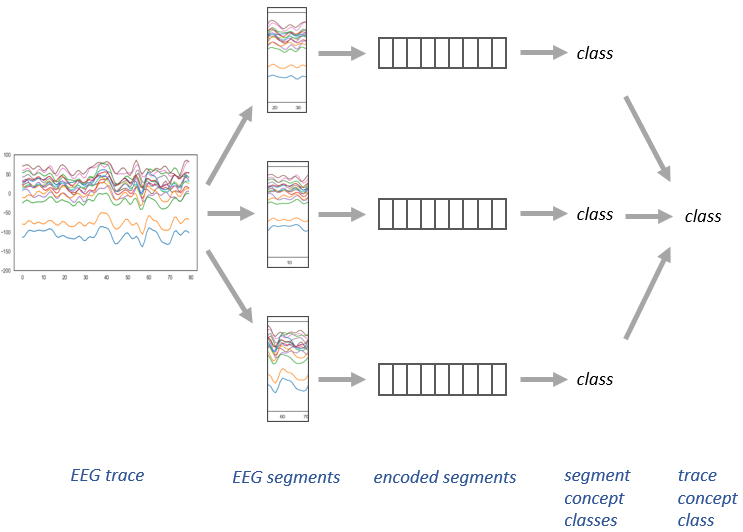}
\end{center}
\caption{The trace classification process.  The input EEG trace is segmented 
and then 
each segment is encoded and classified.  The resulting concept classes are
combined to identify a concept class for the trace as a whole.}
\label{fig:system-operation}
\end{figure}

Predicting a concept class from an EEG trace involves the four steps shown 
from left to right in Fig. \ref{fig:system-operation}.
First, the input trace is segmented using a sliding window, yielding a 
collection of segments.  Each segment is then encoded by mapping it to
a low-dimensional space.  The concept class of each segment is then predicted.  
In particular, for each segment, the probability of each concept class 
is computed.  Finally, the predicted class of the trace as a whole is 
computed from the predicted classes of the trace's segments.

This method, in which the encoding of an input is 
performed independently from downstream tasks such as classification,
is called {\em representation learning} \cite{bengio2013representation}.

\begin{figure}[ht]
\begin{center}
\includegraphics[scale=.46]{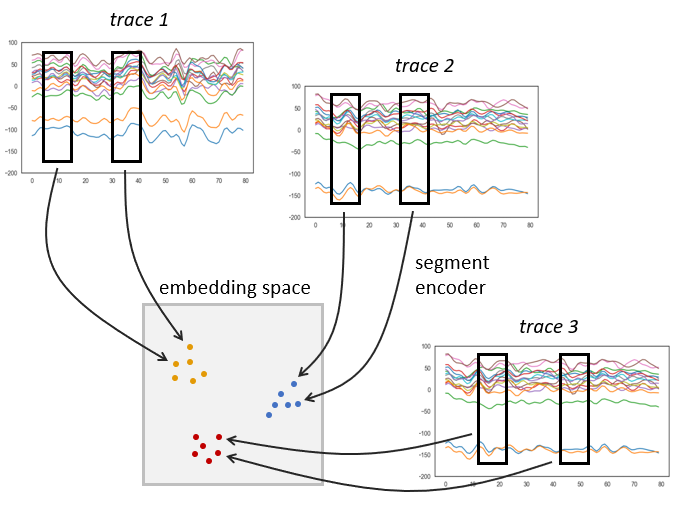}
\end{center}
\caption{The embedding of segments of EEG traces into a lower-dimensional space.}
\label{fig:embedding-space}
\end{figure}

Fig. \ref{fig:embedding-space} illustrates the encoding of the segments 
of a trace to the lower-dimensional embedding space.  
In the figure, the points in 
the embedding space are colored according to the concept associated with 
each trace.  
The segment encoder is trained using segments from a set of training traces;
its goal is to map segments from the 
same trace to nearby points in the embedding space, and segments from distinct traces to distant points.  (The figure is simplified in that it does 
not show the preprocessing
that is applied to segments before they are mapped to the embedding space.)

There are several benefits to using representation learning in segment classification.
First, segment classification can be performed even for classes never seen
during the training of the segment encoder.  Second, the learned representation
of segments can be used to solve problems other than segment classification.
Third, one can hope to gain an understanding of the important features of 
EEG segments by studying the learned embedding space.

The concept class of a trace is then predicted by combining the predicted
classes of the segments in the trace.  For each class, the predicted 
probabilities of the class across all segments are summed and then normalized to obtain a probability distribution over concepts.

The main components of our system are the {\em segment encoder}, which
maps segments to their encodings, the {\em segment classifier}, which
predicts the concept class of an encoded segment, and the {\em trace classifier},
which predicts the concept class of a trace as a whole.  In the following 
sections these system components are described in detail.

\begin{figure}[ht]
\begin{center}
\includegraphics[scale=.6]{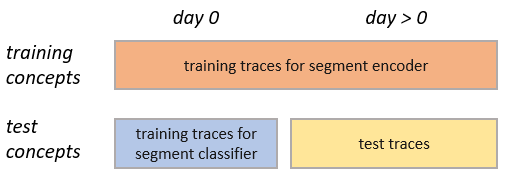}
\end{center}
\caption{Organization of the system's training and test sets.}
\label{fig:training-test-data-2}
\end{figure}

Because the brain activity associated with the recall of a concept changes
over time, special attention is paid to how EEG traces recorded on different
days are used to train and evaluate the system 
(see Fig. \ref{fig:training-test-data-2}).
Concepts are first split into a set of training concepts and a set of test 
concepts.  In training the segment encoder, for each training concept, 
traces from day 0 and later days are used.  In this way, the segment 
encoder can learn what is invariant in the EEG data recorded when 
recalling at multiple points in time.
In training the segment classifier, traces from only day 0 are used.  
In testing, trace concept predictions are made for traces recorded after day 0.  

\subsection{Data segmentation}
\label{subsec:data-segmentation}

The preprocessed EEG traces are segmented using a window size of {\SegLen} samples
with an offset (or ``stride'') of {\SegOffset} samples, giving a segment length of 800 ms.
In segmenting the traces we are relying on the stationarity of EEG data, at least
insofar as concept recall is concerned.  Whether EEG data should be regarded
as stationary is the subject of disagreement in the literature
\cite{FLORIAN1995, Zoldi2000}. 

Each segment is then independently preprocessed using the following sequence
of operations.

\textbf{Trend removal}.  A heuristic method is used to detect trend in
segments.  For each channel in a segment, the trend of the channel is
computed as the absolute difference between the means of the first and 
second halves of the segment. The trend for the segment as a whole is 
then computed as the maximum of channel trends. The 5\% of the segments 
for a concept that most display trend are dropped.  

\textbf{Outlier removal}.  
An anomaly detection algorithm is used to identify the most anomalous
segments of each concept, which are then dropped.
For the purpose of anomaly detection, each segment is mapped to a
feature vector by taking the standard deviation of each channel
separately.  These feature vectors are then provided as input to
the isolation forest anomaly detection algorithm \cite{liu2008}.
The output scores from the algorithm are used to rank the segments,
and the 5\% of the segments deemed most anomalous are dropped. 

\textbf{Normalization}.  Finally, each segment is normalized by applying 
z-score normalization independently to each channel of the segment.

\subsection{The segment encoder}

The segment encoder maps EEG segments to encodings in the embedding space.

\subsubsection{Neural architecture}

\begin{figure}[ht]
\begin{center}
\includegraphics[scale=.43]{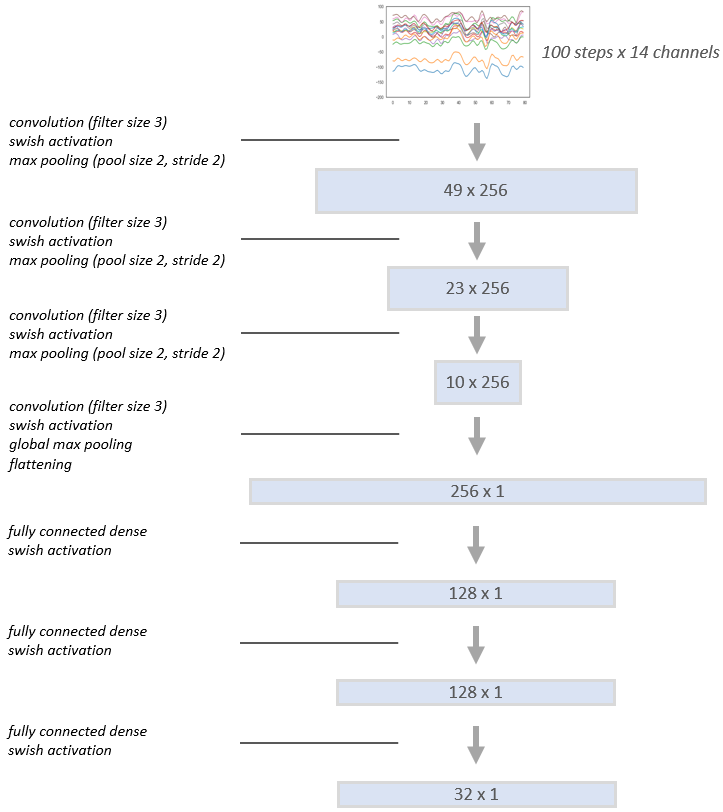}
\end{center}
\caption{Architecture of the segment encoder.}
\label{fig:segment-encoder}
\end{figure}

Fig. \ref{fig:segment-encoder} shows the architecture of the segment encoder.
Processing stages are shown on the left; the dimensions of the data between
stages are shown in the blue boxes.  The encoder input is an EEG segment.
In the first stage of processing, convolution and max pooling are applied.
The convolution operation has {\NumFilters} filters of size {\FilterSize}.  Conceptually, each filter is passed over all {\NumChan} channels of segment 
data, producing a single new time series of about the same length as the 
input segment.  Combining the output from each of the filters yields a 
time series of {\NumFilters} channels.  The output of the convolution 
operation is passed to a max pooling operation of size 2 and stride 2, 
which has the effect of reducing the number of steps in the time series.

After several convolution and pooling blocks are applied, a convolution
operation is followed by global max pooling, which outputs the maximum value
found in each channel, yielding a time series of length 1.  A flattening
operation then transforms the time series into a vector of length
{\NumFilters}.  Informally, these stages of convolution and pooling operations
extract {\NumFilters} features from the input segment.
The use of global max pooling is consistent with our assumption that the
EEG data is roughly stationary; after global max pooling the temporal
dimension of the input data is absent.

The vector of length {\NumFilters} is then processed by several fully-connected
neural layers, resulting in a final encoding of 
size {\EncodingSize}.  Overall, the encoder maps a segment from an input
space of {\SegLen} $\times$ {\NumChan} dimensions into embedding space of
just {\EncodingSize} dimensions.  The neural encoding model itself has 
{\ModelSize} parameters.  The values of the parameters are determined by the training process.

\subsubsection{Loss function}
\label{subsec:loss-fun}

The loss function used to train the segment encoder is supervised 
contrastive loss \cite{khosla2020supervised}.  With this function,
the loss value for a batch of training examples is low if the segments 
associated with the same concepts are nearby in the encoding space, and 
the segments associated with different concepts are widely separated 
in the encoding space.  In other words, the goal of supervised contrastive 
loss is an encoding space in which the segments associated with a concept 
form a cluster in the encoding space.

Khosla {\it et al} (see \cite{khosla2020supervised}, equation 2) define 
the supervised contrastive loss ${\cal L}$ for a set of embeddings as 
shown below.  The set of 
embeddings and corresponding labels is $\{ (z_i, y_i) \mid i \in I\}$
(where $I$ is an index set), $A(i)$ is the set of all indexes
in $I$ except $i$, and $P(i) = \{ j \in A(i) \mid y_j = y_i \}$ is the set of indexes
of embeddings with the same label as $z_i$, but not including $i$.
The definition is parameterized by variable $\tau$, the ``temperature" parameter.
\[
{\cal L} = \sum_{i \in I} \frac{-1}{|P(i)|} 
               \sum_{p \in P(i)} \log \frac{\exp((z_i \cdot z_p)/\tau)}
                                     {\sum_{a \in A(i)} \exp((z_i \cdot z_a)/\tau)}
\]

The definition assumes the embeddings have been $L2$-normalized.
In other words, the length of 
each embedding is 1 when the length is measured using Euclidean distance.
Because the embeddings are $L2$-normalized, the dot product of two embeddings
equals the cosine of the angle between them; the ``cosine distance".

Intuitively, the definition states that the loss for an embedding is based on
the average similarity of the embedding with other embeddings of the same class.
Operationally, the loss for a single embedding $z_i$ can be computed as follows.
First, use the dot product to compute the similarity of $z_i$ and every other
embedding $z_a$.  Second, normalize the similarity values by dividing them 
by their sum. Third, compute the log of each normalized similarity.  
Finally, get the average of these log similarity values for embeddings of 
the same class as $z_i$, and negate it.

Supervised contrastive loss was originally defined \cite{khosla2020supervised} in 
the context of image classification and a learning framework that
includes not just an encoder but also data augmentation and a ``projection head".
The data augmentation component produces two modified images from a single input
image.  The projection head is a feedforward neural network with only one or
two layers.  In training, the output of the encoder is fed to the projection head.  
After training, the projection head is not used.

In our model, the projection head is a feedforward neural network with two layers, 
each of \ProjHeadSize{} neurons.  Data augmentation in the style of
\cite{khosla2020supervised} is not used. 
We use a variant \cite{Salama2020} of the definition of supervised contrastive
loss shown above.  In the variant, the definitions of $P(i)$ and $A(i)$ are 
modified so that both $P(i)$ and $A(i)$ are extended to include $i$ itself.  
In other words, the similarity of $i$ with itself is included in the loss value. 
We use a temperature parameter value of 0.1.

\subsubsection{Training}

Training of the segment encoder proceeds in steps.  In each step, the encoder
is given a set of segments, called a {\em batch}.  In cost functions used with 
one-shot learning, system performance can depend strongly on how training 
examples are selected for each batch.  In representation learning with supervised contrastive loss, experiments suggest that accuracy improves with larger batch size \cite{khosla2020supervised}.  In our system, each batch contains {\BatchSize} 
segments for each training concept, giving a batch size of $\NumTrainingConcepts 
\times \BatchSize$ segments.  Batches are generated on-the-fly with a batch 
generation function.

To reduce model overfitting, zero-centered
Gaussian noise is added to each segment, with variance controlled by a system tuning
parameter.  Training proceeds for {\NumEpochs} epochs, each having {\StepsPerEpoch} 
steps. An rmsprop \cite{Hinton-rmsprop} optimizer is used.

\begin{figure}[ht]
\begin{center}
\includegraphics[scale=.45]{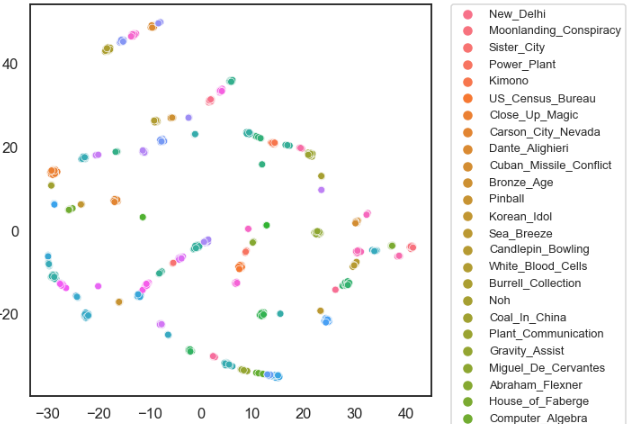}
\end{center}
\caption{Embeddings of training segments after mapping to two dimensions.}
\label{fig:training-embeddings}
\end{figure}

Fig. \ref{fig:training-embeddings} shows the embedding of about 1000 random segments 
after training the segment encoder.  The embeddings, of length \EncodingSize, have 
been mapped to 2-dimensional space using t-distributed stochastic neighbor 
embedding (tSNE) \cite{van2008}.  Each point represents a segment and is 
colored according to the Wikipedia page concept it is associated with.  The legend 
lists only some of the concepts contained in the training data set.

\begin{figure}[ht]
\begin{center}
\includegraphics[scale=.45]{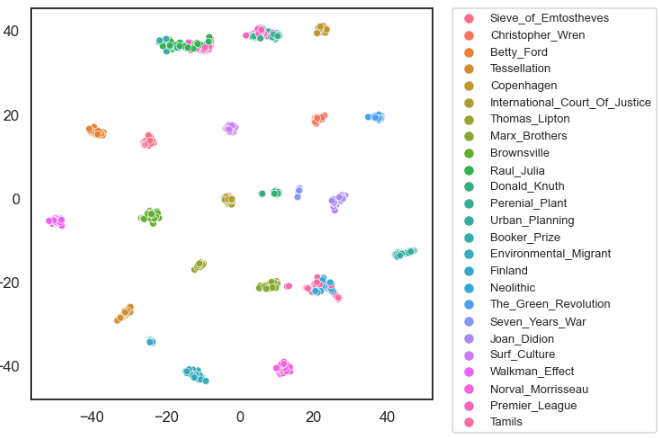}
\end{center}
\caption{Embeddings of test segments after mapping to two dimensions.}
\label{fig:test-embeddings}
\end{figure}

Fig. \ref{fig:test-embeddings} shows the embeddings of segments associated with
concept classes not seen during the training of the segment encoder.  As in Fig. \ref{fig:training-embeddings}, the embeddings have been mapped to 2-dimensional
space using tSNE.  Even though these concept classes were not seen during training,
the segments cluster well in the embedding space according to the concepts.

\subsection{The segment classifier}

The segment classifier takes an encoded EEG segment as input and outputs a concept 
class.  Any machine learning algorithm can be used to create the segment classifier.
We use a k-nearest-neighbors (KNN) classifier with {\KNearest} nearest neighbors,
and Euclidean distance as the distance function.

Our KNN classification method supports predictions in the 
form of a probability distribution over the concept classes.  In other words, the
prediction function takes as input an EEG segment and outputs a
probability distribution over concept classes.
KNN classification is a form of instance-based learning, in which the 
training data itself serves as the model, and therefore the training process
involves only storing the training data.

\subsection{The trace classifier}
\label{sec:trace-classifier}

The trace classifier takes an EEG trace as input and outputs a probability 
distribution over concept classes.  It is an ensemble classifier that
predicts the concept class of a trace from the predictions of concept classes
in the segments of the trace.

The trace classifier first preprocesses and segments the input trace
(as described in Section \ref{subsec:data-segmentation}),
then applies the segment classifier to classify each segment.  
Because the segment classifier's output for each input segment is
a probability distribution over concept classes, a ``soft" method is
used to combine the segment predictions to form a prediction for the trace as a whole.
For every concept class, the predicted probabilities for the class across
all segments are summed.  The resulting vector of sums is then normalized
to obtain a probability distribution over the concept classes.
The trace classifier needs no training of its own.

\section{Results}
\label{sec:results}

Here we describe our experimental setup and experimental results.
Our experiments attempt to answer two broad questions.
First, what is the performance of our system in predicting the
concept class of an EEG trace?  Second, what can our system tell
us about the brain's memory function? 

\subsection{Experimental setup}

To test our system we perform a number of experimental trials, where
a single trial consists of the following steps:
(Recall that our data set contains EEG traces for
{\NumConcepts} concepts.)

\begin{enumerate}
    \item  A test/train split is performed.  \NumTestConcepts{} concepts are 
    randomly selected, and segments for these concepts are put in a test set.  
    Segments for the remaining concepts are put in a training set.

    \item The segment encoder is trained on batches from the training set.
    
    \item A KNN classifier is created using the day 0 segments from the
    test set as training data.
    
    \item Concept predictions are made by the KNN classifier on day 1 segments
    from the test set.  Top-1, top-2, and top-3 accuracy values are computed
    using the concept labels of the test set.
\end{enumerate}
    
Running multiple experimental trials on the system yields a set of accuracy
values with high variance.  One source of the variance is the random 
initialization of weights in the training of the source encoder.  
Another is the random split of the concepts into training and test sets.
Also, every trial contains only \NumTestConcepts{} concepts, so the
resulting accuracy for the trial will be a value in $\{0.0, 0.05, 0.10, \ldots, 1.0\}$.

Variance in experimental results makes the evaluation of system performance
difficult, which in turn makes system tuning difficult.  To address this
problem, \NumTrials{} trials are run, and then the bootstrap is used to compute
a confidence interval for accuracy values.   

This approach, in which multiple trials are run, each with its own randomly-chosen
test/train split, is a variant of cross-validation.  In cross-validation, a
set of training examples is randomly split into $n$ disjoint groups, called
{\em folds}, and then a training and test process is performed $n$ times.
Each time, a distinct fold is used as the test set, and the remaining folds
are combined for use as the training set.  We cannot use simple cross-validation because we want training and test sets to contain segments of
different concepts.  We do not simply split the concepts into $n$ folds because
then larger values of $n$ lead to folds containing a small number of concepts,
which makes the test results unreliable.  By splitting the concepts into
random groups instead of $n$ disjoint folds, we can separately control
the number of trials and the number of concepts represented in a test set.

In reporting test results, we provide the mean top-1 and top-3 accuracy 
values across all trials in a test. 

\subsection{Trace and segment classification accuracy}
\label{sec:trace-accuracy}

{\bf Trace classification accuracy.}
The most important performance statistic for our system is the trace
classification accuracy: the accuracy in predicting a concept given an
EEG trace of the subject thinking of the concept.

Our system achieves a mean top-1 validation accuracy of ${\TopOne}\%$, and
a top-3 validation accuracy of ${\TopThree}\%$.
Our test set contains \NumTestConcepts{} concepts, with 1 trace for each
concept, so our top-1 accuracy of $\TopOne{}\%$ can be compared to a baseline of 0.04 that could be achieved
by chance.  Similarly, our top-3 accuracy of $\TopThree{}\%$ can be compared
to a baseline of 0.12.
These top-1 and top-3 accuracy values were achieved on concepts that were
not used in the training of the segment encoder.

\begin{figure}
\begin{center}
\includegraphics[scale=.62]{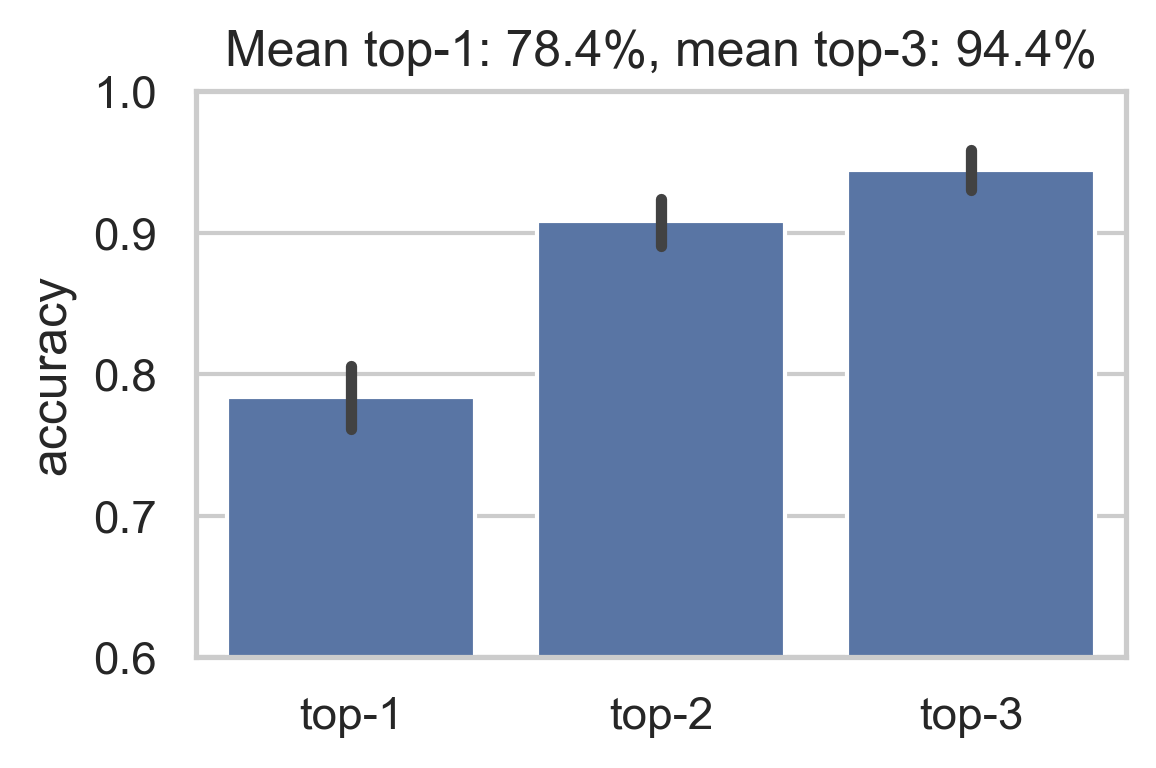}
\end{center}
\caption{System trace classification accuracy.}
\label{fig:accuracy}
\end{figure}

Figure \ref{fig:accuracy} shows our overall performance results, based
on 50 trials.  The error bars in the figure represent a 
$95\%$ confidence interval obtained by using the bootstrap on the
accuracy values from individual trials.

{\bf Segment classification accuracy.}
The trace classifier uses predictions on many individual segments
to make its own prediction.  What accuracy is achieved by the
segment classifier on a single segment?
To evaluate segment classification performance,
50 trials were performed using our baseline hyperparameter settings.
In each trial, classification was performed on all of the
day 1 test segments.  In all, about 1.06 million segments were classified;
on average about 10,300 classifications per concept.

\begin{figure}
\begin{center}
\includegraphics[scale=.55]{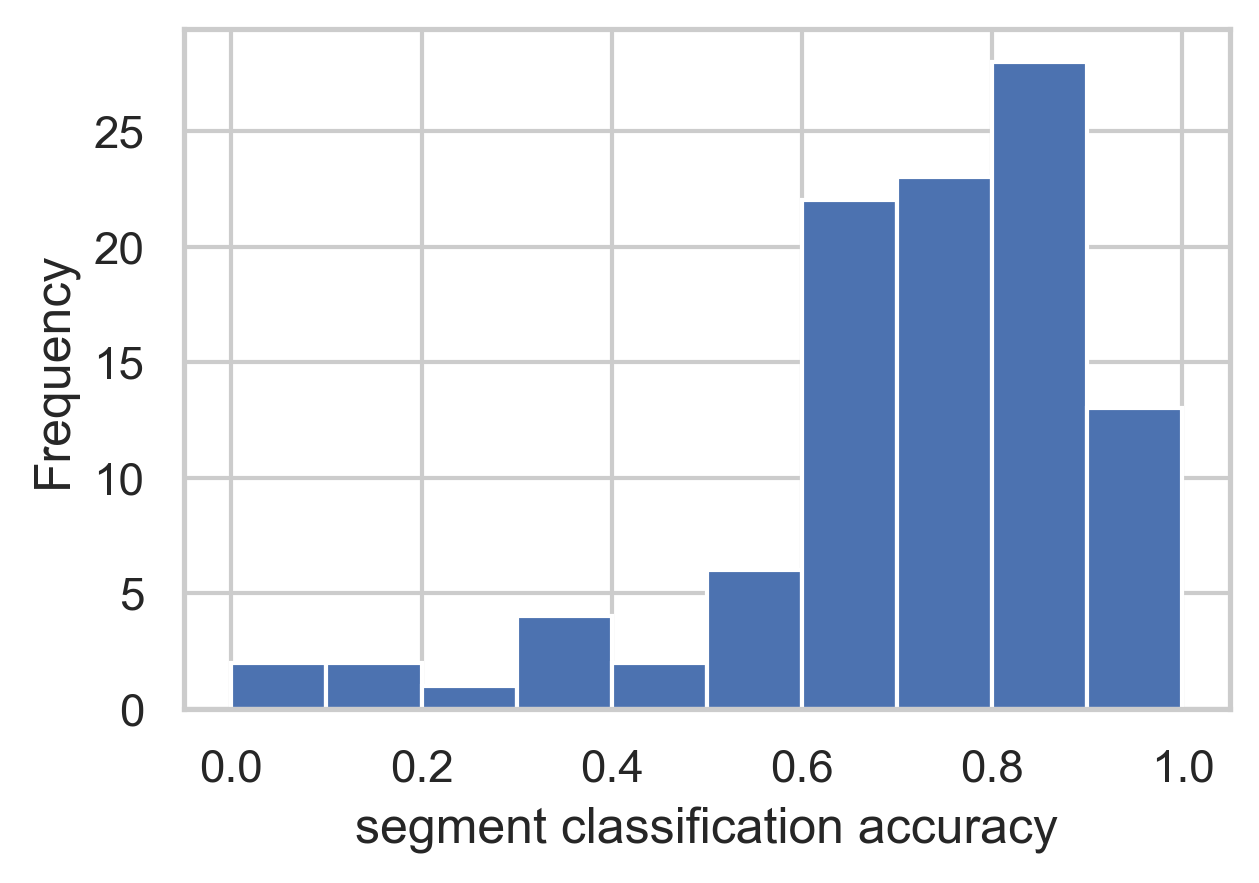}
\end{center}
\caption{Histogram of per-concept classification accuracy.}
\label{fig:seg-accuracy}
\end{figure}

An overall segment classification accuracy of 0.725 was obtained,
versus 0.04 by chance.  However, accuracy varied significantly
by concept.  Figure \ref{fig:seg-accuracy} shows the distribution
of the per-concept accuracy values.  For
some concepts, the segment classification accuracy is above 90\%, while
for others, it is below 10\%.

Per-concept segment classification accuracy varies not just across
concepts, but also across trials.  In each
trial a different, random split of concepts into training and test
concepts is used.  The train/test split controls which concepts
are used to train the segment encoder, and the set of test
concepts that must be distinguished in testing.

\begin{figure}
\begin{center}
\includegraphics[scale=.47]{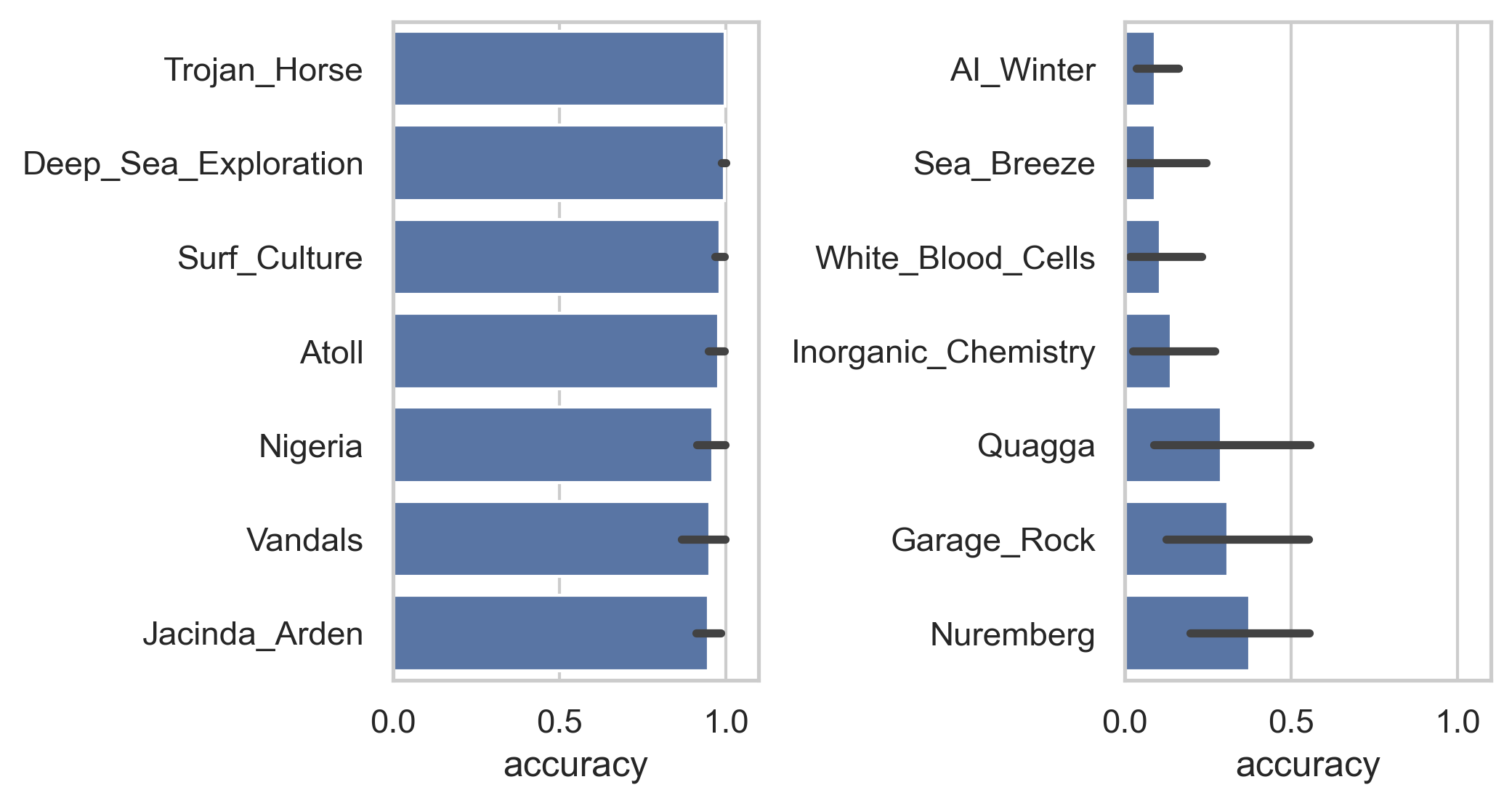}
\end{center}
\caption{Average concept classification accuracy for concepts with high average accuracy (left) and low average accuracy (right).}
\label{fig:best-worst-concept-accuracy}
\end{figure}

Fig. \ref{fig:best-worst-concept-accuracy} shows how
per-concept segment classification accuracy varies across
trials.  The plot on the left shows the mean accuracy across
trials for the concepts with
the highest mean accuracy, while the plot on the right does the
same for the concepts with the lowest mean accuracy.
95\% confidence intervals obtained by bootstrapping are also shown.

The figure shows that some concepts, such as ``Algebraic Structure", achieve near-perfect segment classification accuracy in every trial in which they appear, while others, like ``Bohemianism", achieve very low accuracy across trials.  The segment classification accuracy for concepts like ``Kimono" and ``Inorganic Chemistry" shows high variance across trials -- and thus depends on the concepts used for training.

\subsection{EEG node and frequency importance}
\label{subsec:node-freq-importance}

Here we look at the predictive power of individual EEG channels and
EEG frequency bands.  For example, we look at the trace prediction
accuracy that can be obtained if the EEG data only contains data from
the F3 node.  

\begin{figure}
\begin{center}
\includegraphics[scale=.35]{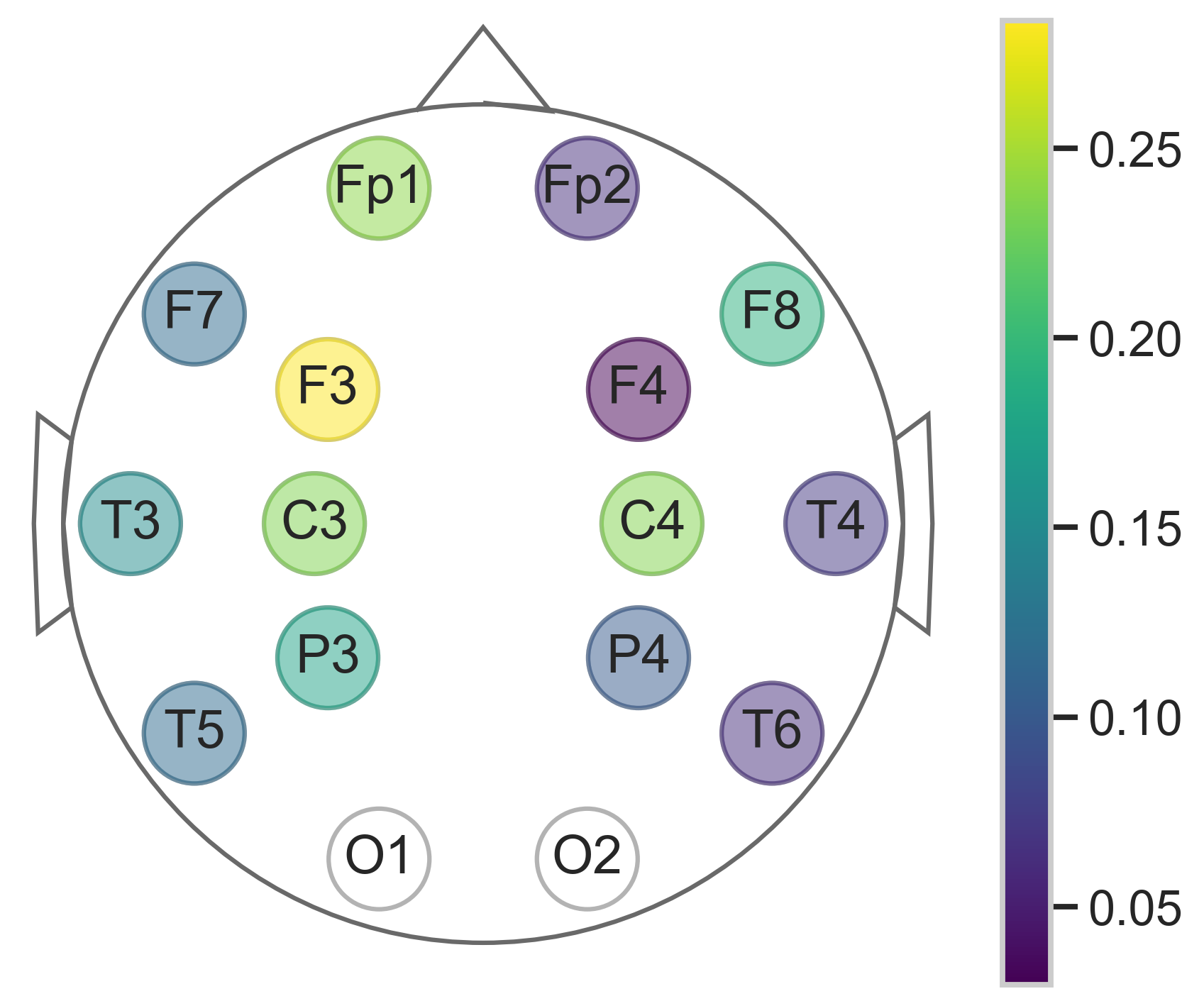}
\end{center}
\caption{Average trace classification accuracy using data from individual EEG nodes.}
\label{fig:single-channel-importance}
\end{figure}

{\bf Predictive power of EEG channels.}
Fig. \ref{fig:single-channel-importance} shows the average trace
prediction accuracy obtained when trained and tested with data
from only a single EEG channel.  Here, ``average" means the average
of the top-1, top-2, and top-3 trace accuracy values.
For example, a version of our data set was created containing only 
data from the F3 node. No tuning of our system was performed to 
optimize its accuracy with this data set; hence, the accuracy values 
shown in the figure are conservative.  

The figure shows that the F3 node provides the highest predictive
power of any node taken alone, with an average accuracy of about 0.28.
Next in predictive power are the Fp1, C4, and C3 nodes.  The nodes
with the lowest predictive power are the F4, Fp2, and T6 nodes.
The occipital nodes are left uncolored because occipital data is
removed during preprocessing.

While EEG may not have the spatial resolution offered by fMRI, PET, 
or ECoG, the predictive power of specific EEG channels correlates with 
known brain functions and their regions within the brain. Channels C3 and F3 
cover the superior frontal gyrus. Activation in the left superior 
frontal gyrus has been linked to semantic categorization, the process of determining the class or group of a concept \cite{devlin2002}. Channel Fp1 covers the left prefrontal cortex, which has long been linked to semantic processing, including semantic working memory \cite{Gabrieli1998}.

\begin{figure}
\begin{center}
\includegraphics[scale=.35]{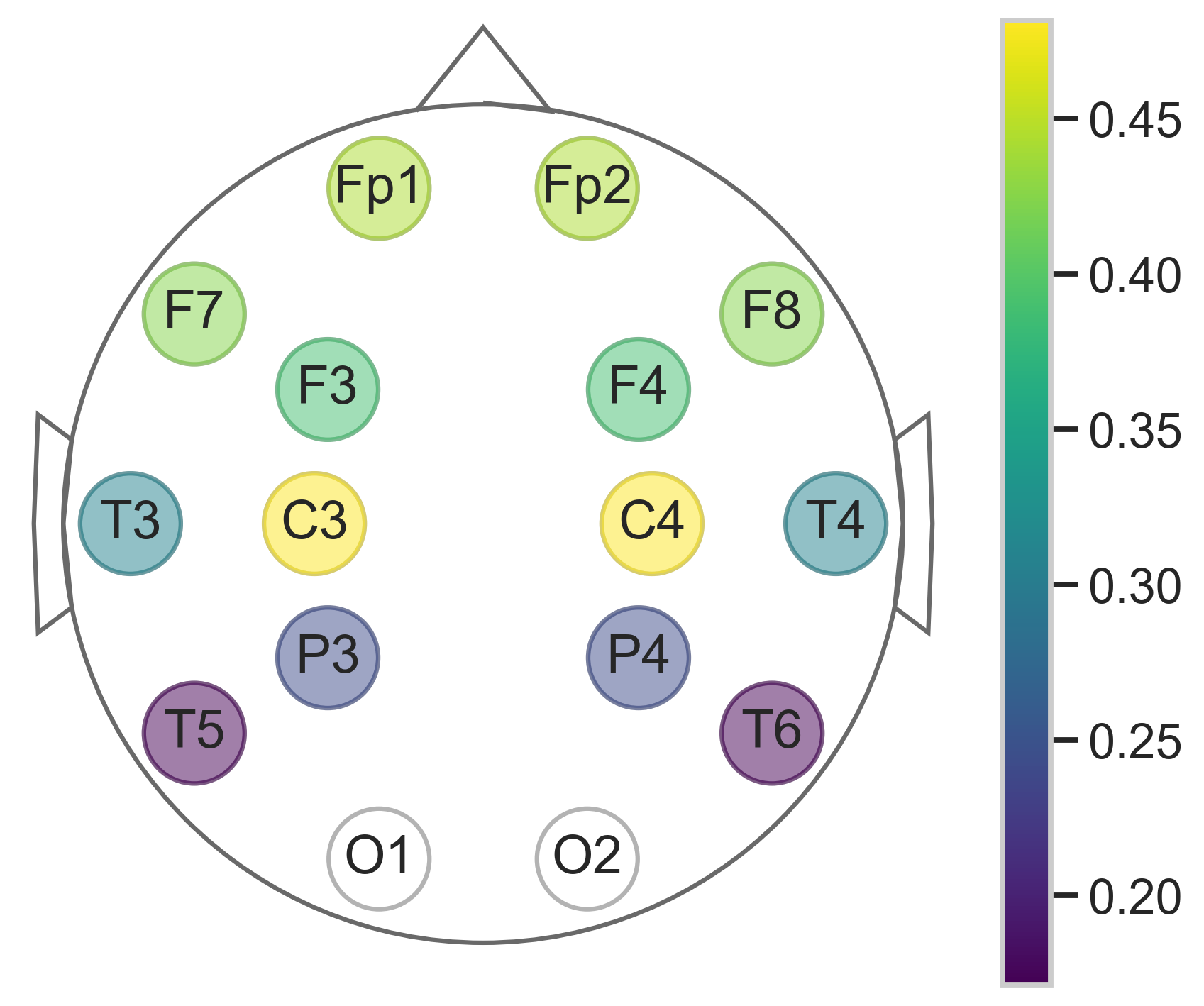}
\end{center}
\caption{Average trace classification accuracy using data from pairs of EEG nodes.}
\label{fig:channel-pair-importance}
\end{figure}

Fig. \ref{fig:channel-pair-importance} shows the average trace prediction
accuracy for data sets from bilateral pairs of EEG nodes.
The highest average trace accuracy of about 0.48 is obtained using 
the central nodes.  
We hypothesize this is due to their central location combined with low spatial resolution of the EEG. The nodes, located in proximity to the frontal cortex, encompass the angular and supramarginal gyrus, which have been linked to sentence processing \cite{pallier2011}, semantic memory \cite{binder2005distinct}, and various phonological processes \cite{Oberhuber2016}. Channels P3 and P4 scored half as well (0.242) as the central nodes. We speculate this reflects incomplete information of cortical reinstatement activity during a concept's retrieval.

The prefrontal (Fp1 and Fp2) and lateral prefrontal (F7 and F8) pairs show the second and third highest average trace accuracy, respectively. The Fp1 and Fp2 nodes cover the prefrontal cortex which is responsible for many functions; most relevant to this paper are goal direction and top-down memory retrieval \cite{Miller2001}. The F7 and F8 nodes are expected to contain information from the medial and inferior frontal gyri, including Broca's area. These areas are known to play key roles in semantic retrieval and working memory \cite{LIAKAKIS2011}. 

\begin{figure}
\begin{center}
\includegraphics[scale=.35]{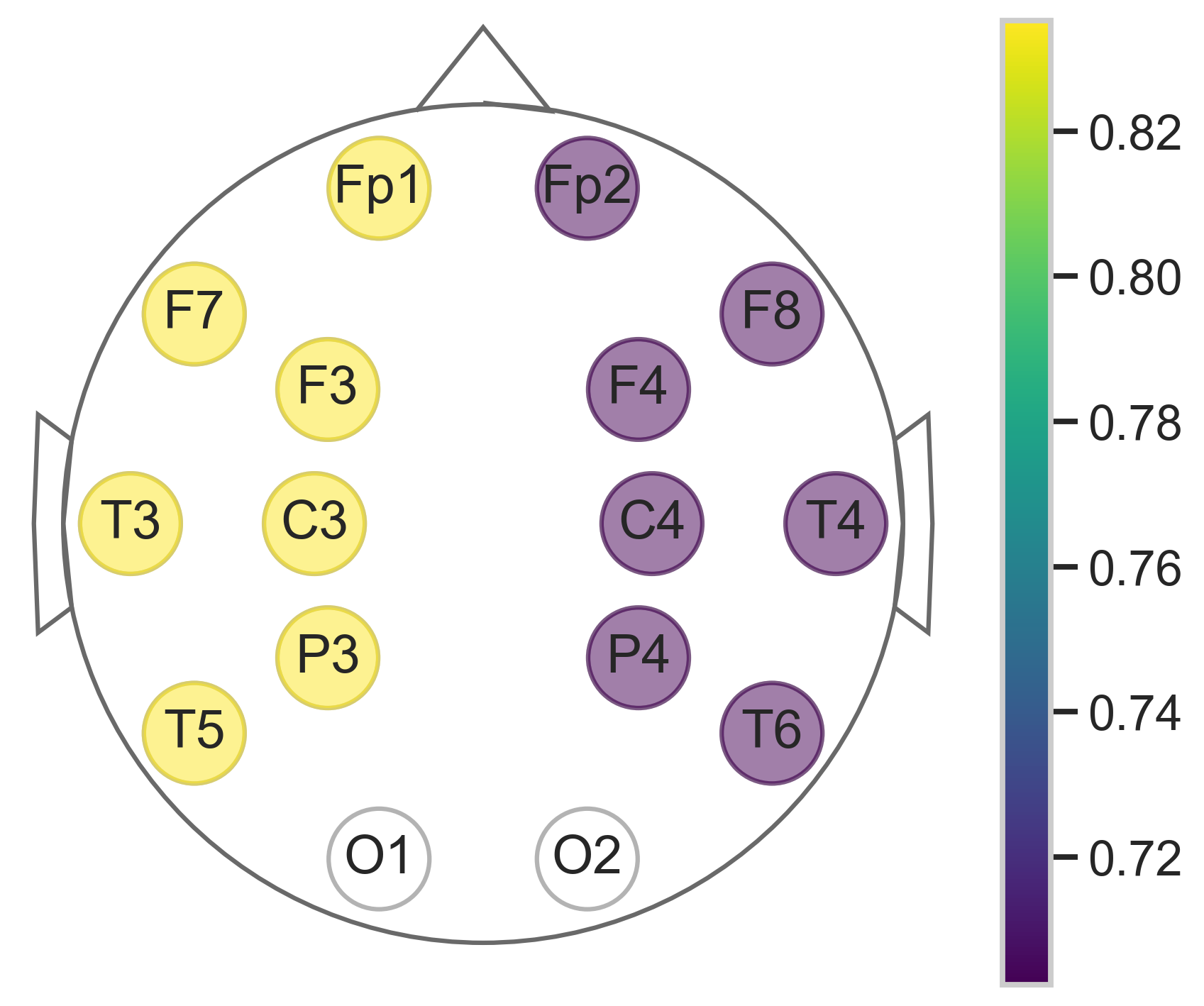}
\end{center}
\caption{Average trace classification accuracy using data from nodes of the left and right hemisphere.}
\label{fig:hemisphere-importance}
\end{figure}

Fig. \ref{fig:hemisphere-importance} shows the predictive power
of the nodes in the left and right hemispheres taken on their own.
The nodes of the left hemisphere have higher predictive power than
the nodes of the right.  The average accuracy for the left is
about 13\% better than the average accuracy for the right. These results are consistent with research showing asymmetrical hemisphere activation during semantic processes, including memory \cite{REILLY2015, Mummery1991, MARTIN2001, Golby2001}. 

\begin{figure}
\begin{center}
\includegraphics[scale=.6]{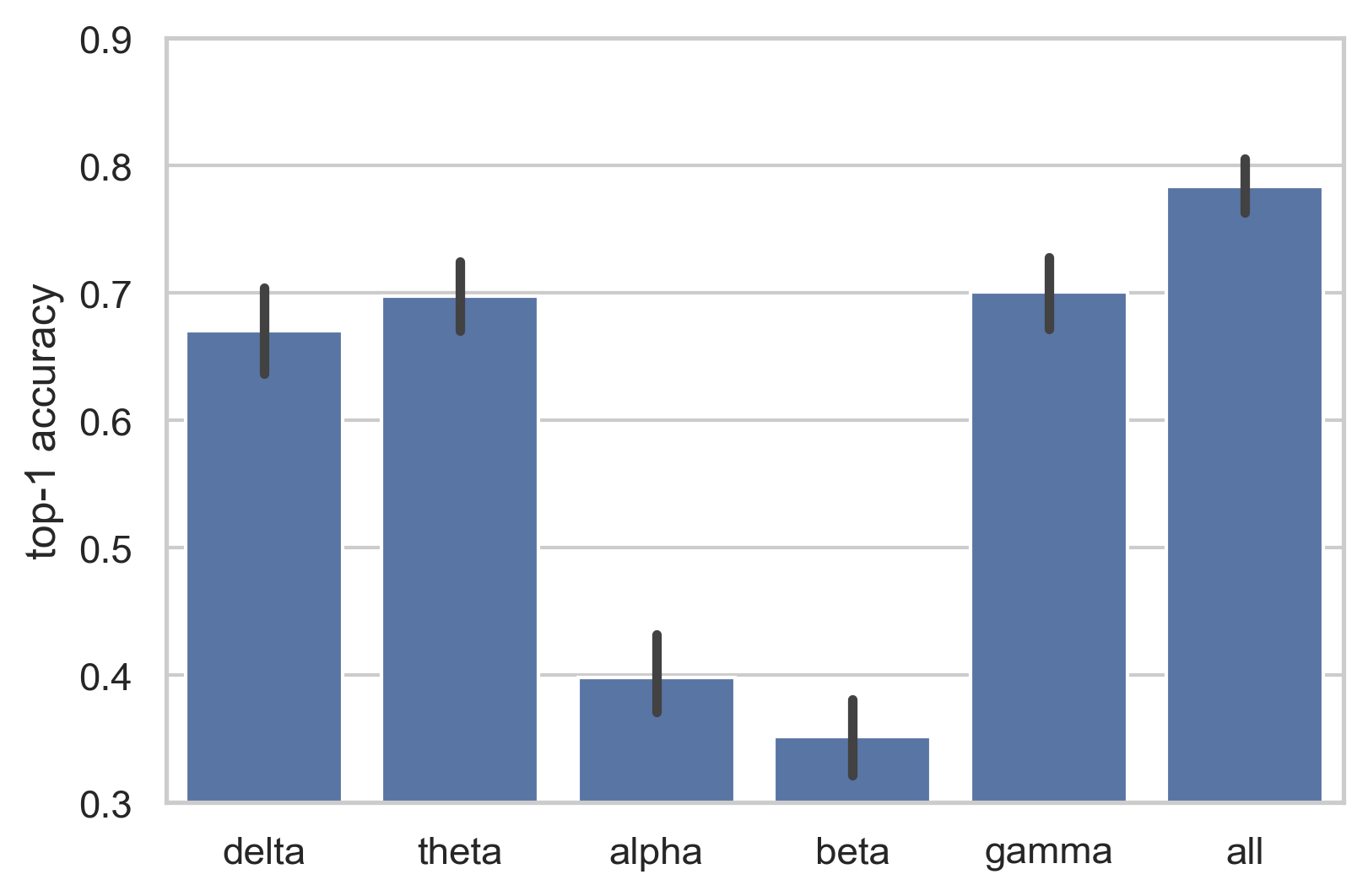}
\end{center}
\caption{Top-1 trace classification accuracy by EEG frequency band.}
\label{fig:frequency-importance}
\end{figure}

{\bf Predictive power of EEG frequency bands.}
Figure \ref{fig:frequency-importance} shows the top-1 concept prediction 
accuracy for EEG data from single frequency 
bands.  At least 20 trials were performed for each 
frequency band.  Also shown is the concept prediction accuracy
when all of the delta through gamma frequency bands are used.
Trace classification accuracy as a function of frequency band serves 
two main roles: it provides validation of the cognitive functions 
being learned by our model and gives insight into the frequencies 
that contain information on the content of semantic memories. 

Several studies have investigated the connection between memory and 
frequency bands. Memory-related coherence, a measure of synchronization 
between brain regions, has been seen within the lower frequencies (1-19 Hz) \cite{Fell2003}. Delta and theta waves have also been linked to working memory and attention \cite{KNYAZEV2012,Fernandez2002}. Most significantly, gamma and theta 
bands, which produced the two highest accuracies among individual bands, show the most activity during 
semantic memory retrieval \cite{Hanouneh2018}. Decrease in alpha and beta oscillations is also well known to occur during memory tasks from attention to memory retrieval, for an in-depth discussion see \cite{hanslmayr2012}. Working memory, attention, and semantic
retrieval are underlying cognitive functions potentially used 
to decode the content of a memory trace within our system.  

\subsection{Effects of recollection time}

EEG data was collected when the subject recalled a concept just after
learning it, one day after learning it, and three days after
learning it.  In section \ref{sec:trace-accuracy}, we reported
on our model's accuracy in predicting a concept given EEG collected
one day after learning it.  

\begin{figure}
\begin{center}
\includegraphics[scale=.8]{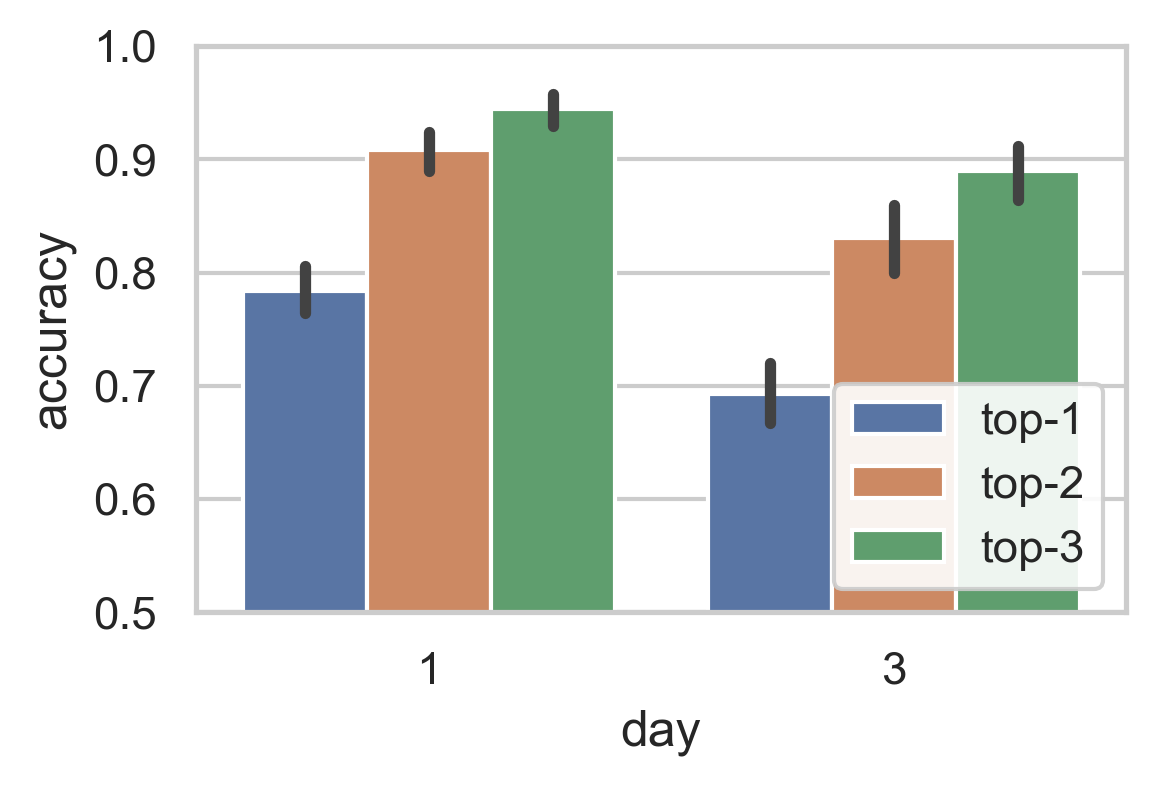}
\end{center}
\caption{Accuracy by day of recollection.}
\label{fig:day1-3-accuracy}
\end{figure}

Figure \ref{fig:day1-3-accuracy} shows top-1, top-2, and top-3 trace prediction 
accuracy for days 1 and 3.  Top-1 accuracy drops from about 0.7 to about 0.55, 
while top-3 accuracy shows a smaller decline, from about 0.95 to about 0.91. 

We speculate the drop in accuracy between days reflects memory consolidation, 
the process by which, over time, memories are reorganized \cite{squire2015}. Observing which changes affect memory prediction could lead to a better 
understanding of memory consolidation and neural plasticity, a use case 
for our system. For example, testing for Alzheimer's disease involves 
expensive MRI equipment to observe brain fluids, spinal taps to extract 
cerebral spinal fluid, or blood sampling and testing for amyloid proteins. 
We speculate that our system could lead to new indicators, such as the 
prediction accuracy delta between days, that signal for further, more 
invasive, testing. Similar applications may also be found in brain trauma, 
aging, and other, memory-related research.   

\section{Neural information retrieval}
\label{sec:app}

In this section we describe an application of neural memory decoding
to information retrieval, and present experimental results pertaining
to whether the application is practical.

The application would work as follows.  A user encounters a document
of interest.  To {\em index} the document, the user briefly recollects
the content of the document while EEG data is collected.  The EEG trace
for the document is stored along with a link to the document.  Later,
to retrieve the document, the user creates a {\em query} by again
recalling the content of the document.  The application then returns
an ordered list of document links, ordered by estimated relevance.
Only previously-indexed documents would appear in the list.
Any kind of document could be used, provided it has a unique identifier,
such as a URL.  The document need not be a text document.  It could be
an image, an audio recording, or a video recording, for example.

Before using the system, the user would need to customize the system
through a training process.  In this process, the user would be
asked to provide EEG data for a collection of short documents.  For each 
document, the user would first read (or otherwise experience) the
document, then record EEG data while imagining it.

This application addresses a pressing real-world problem.
As more information becomes easily available through the internet,
locating a document that one has previously discovered becomes
more difficult.  Solutions such as creating and storing bookmarks rely on
user-created labels, which can be hard to create and provide little
information about a document.  EEG data of the recollection of a
document, on the other hand, serves as an information-rich label.

Fig. \ref{fig:training-test-data-2} can be revisited in the context
of this application, which we call ``neural information retrieval''.
The application would be trained using EEG traces for a set of training
concepts recorded on day 0 and one or more later days.
To index a document (day 0), an EEG trace for the document would be 
recorded and stored.  (Recall that, with a KNN segment classifier,
no computation is needed for training.)
To query a document (day $>$ 0), an EEG trace for the document would
be recorded and then used as input to the trace classifier.
This process is much like the process of our experimental setup, except
that in the application a query could happen at any time after day 0.

Neural information retrieval would not be practical if users would have to
spend too much time collecting EEG data for system training, for indexing
a document, or for searching for a document.  The time to index and search for
documents is most important, as training costs are incurred only once.
We record EEG at {\SampleRate} Hz, so a segment of 100 samples represents 
about 4/5 seconds.  In other words, 10 segments can be recorded in about 8 seconds,
even assuming the segments have no overlap.

\begin{figure}
\begin{center}
\includegraphics[scale=.8]{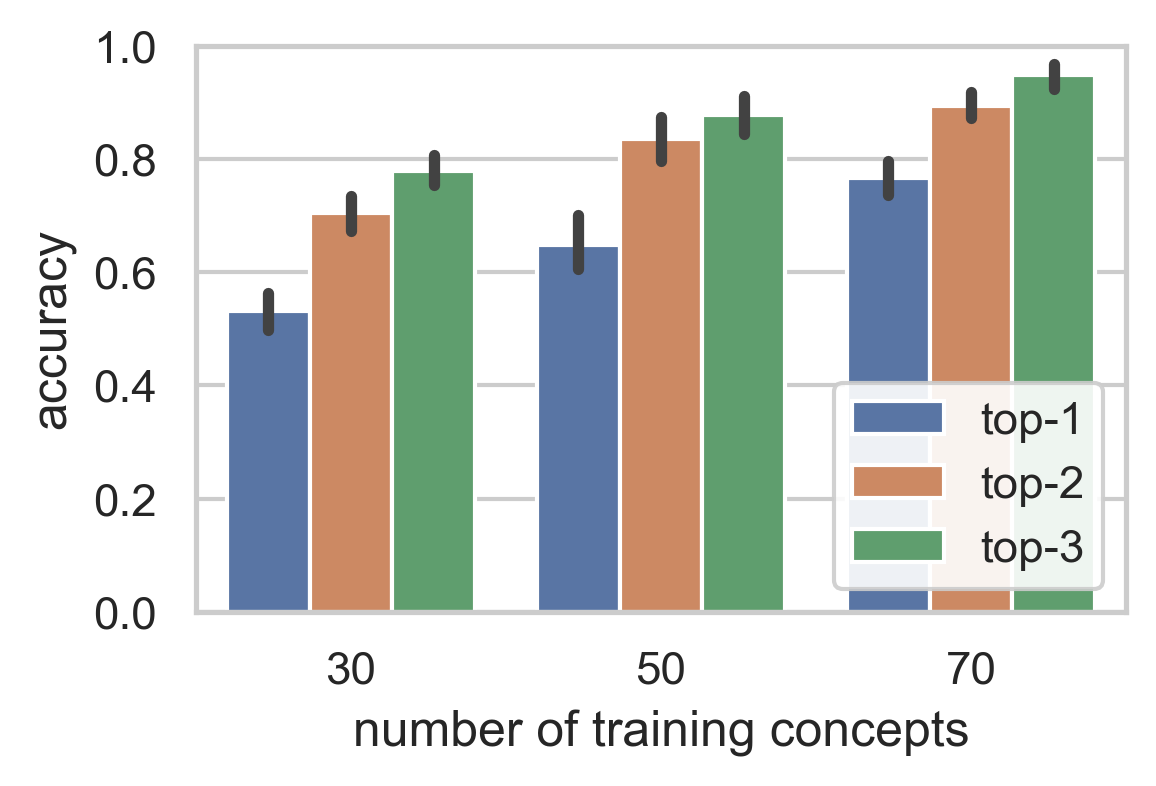}
\end{center}
\caption{Accuracy by number of training concepts.}
\label{fig:accuracy-by-num-train-concepts}
\end{figure}

Fig. \ref{fig:accuracy-by-num-train-concepts} shows trace
classification accuracy as a function of the number of concepts
used to train the system.  Data was collected from 20 trials, and
all segments associated with each concept were used.
Accuracy improves with the number of training concepts, but with 70 
concepts the top accuracy values are about the same as is achieved
with the full training set of \NumTrainingConcepts{}.

\begin{figure}
\begin{center}
\includegraphics[scale=.8]{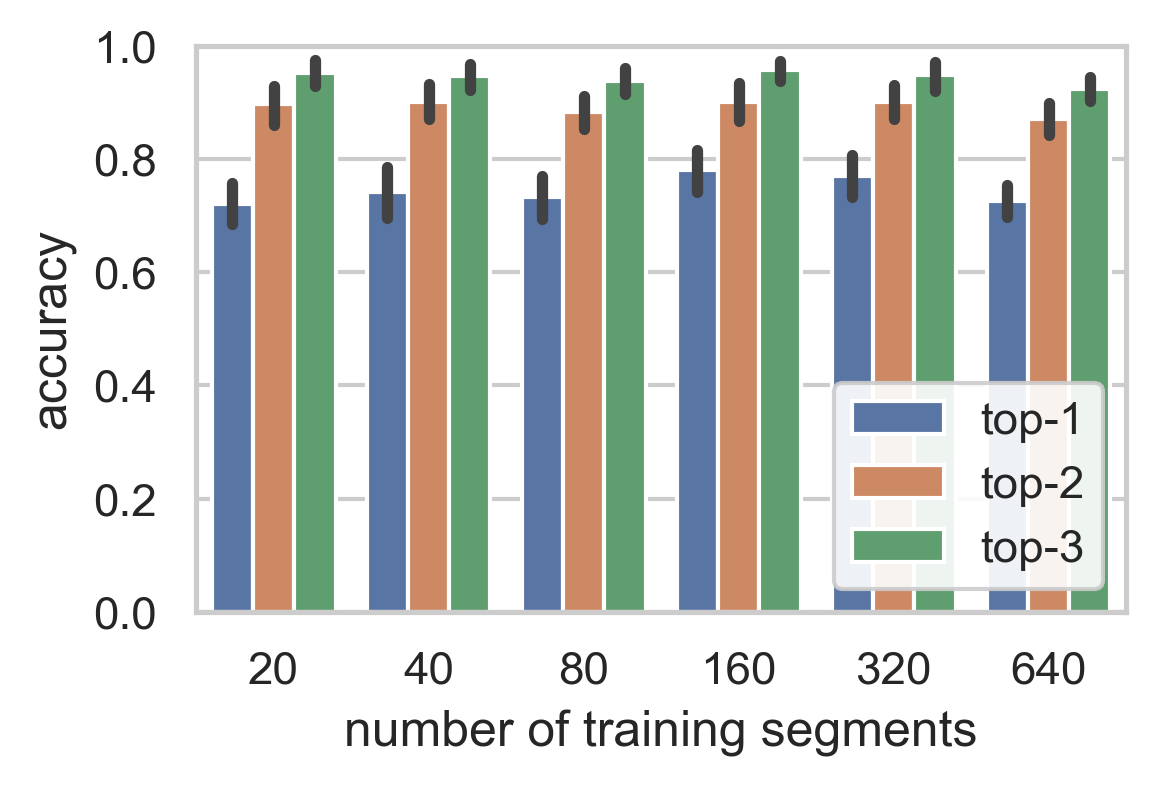}
\end{center}
\caption{Accuracy by number of training segments.}
\label{fig:accuracy-by-num-train-segments}
\end{figure}

Fig. \ref{fig:accuracy-by-num-train-segments} shows trace
classification accuracy as a function of the number of training segments
used per concept.  All training concepts were used.
Notably, only a small improvement is seen as the number of training concepts
is increased from 20 to 320.  No improvement is seen as the number of
segments is increased from 320 to 640.

\begin{figure}
\begin{center}
\includegraphics[scale=.8]{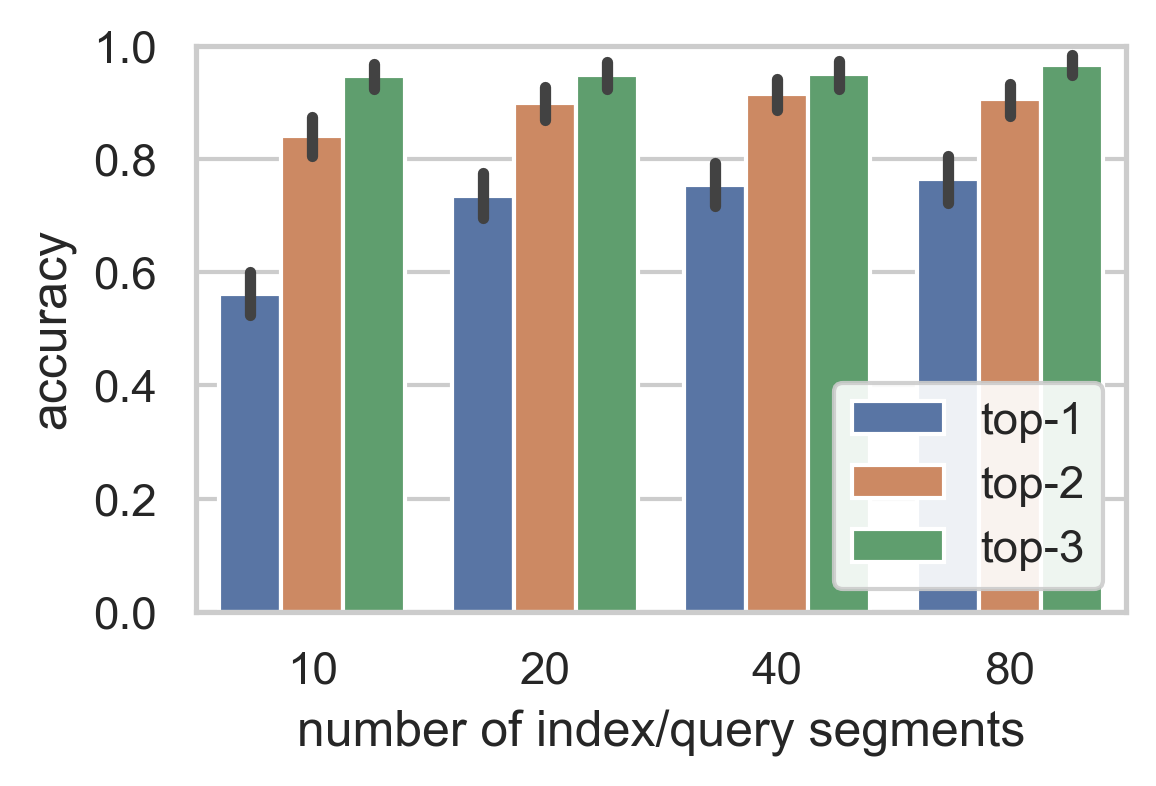}
\end{center}
\caption{Accuracy by number of index/query segments.  All training segments are used.}
\label{fig:accuracy-by-num-index-segments}
\end{figure}

Fig. \ref{fig:accuracy-by-num-index-segments} shows trace
classification accuracy as a function of the number of indexing and
querying segments used.  
All training segments were used, and the 
number of indexing segments is equal to the number of querying segments. 
As the number of segments increases, top-1 accuracy increases, but
there is a smaller improvement in top-3 accuracy.
An average top-3 trace classification accuracy of about 0.94 can be 
obtained with 10 segments, compared to a value of 0.96 when 240 segments are used.  
On the other hand, the top-1 trace classification is about 0.56 with 10
segments, compared to 0.76 for 80 segments.

\section{Conclusions}
\label{sec:conclusion}

We have demonstrated the feasibility of neural memory decoding
with EEG by creating a system that can predict a concept being recalled
with about {\TopOne}\% top-1 accuracy ({\TopOneChance}\% chance) and
about {\TopThree}\% top-3 accuracy ({\TopThreeChance}\% chance).
One benefit of the work is the
insight into memory function that can be obtained by studying how
system performance varies as a function of the EEG frequencies
that are used, the EEG nodes that are used, and the amount
of data that is used.  For example, we found that the single
best EEG channel, in terms of concept prediction accuracy, was
the F3 channel, and that the single best frequency band was the gamma
band.

Another benefit of the work is the potential of the system in
applications.  We described a document retrieval application in
which a user can retrieve a document by simply thinking about it.
Our experiments shed light on whether this application might 
be practical in terms of the time a user would need to record
EEG data when training or using the system.

Our system has serious limitations that need to be addressed, both
for scientific reasons and to understand the potential of neural memory
decoding-based applications.  First, our EEG data comes from just one individual.  
Second, for each concept, we collected data only over a 4-day period.
Would our system perform well if a concept were recalled a month or a year after
being first encountered?  

Regarding future work, from a neuroscience perspective we would like to 
understand the features our convolutional model discovers in EEG data.  
Fortunately, many techniques exist for visualizing 
convolutional neural nets.  For example, one can visualize the 
trained filters, or even superimpose a heat map on an input to see 
which regions of the input are most important to the classification 
process \cite{selvaraju2017grad}. This could lead to new measures of 
memory consolidation or interference during recall. 

Rapid advances are being made in the area of deep learning with neural
nets.  We would like to experiment with newer neural architectures, and
to perform further system tuning to see how much our trace classification
accuracy can be improved.

One can imagine many useful application of neural memory decoding.
We would like to understand the practical issues associated with such applications, such as the amount of training
data needed for a system that can distinguish thousands of concepts,
the amount of indexing and query data needed, and whether
reinforcement learning or other methods could be used to
support applications that improve as they encounter more and more brain data.

The use of bulky, expensive EEG headsets may seem an obstacle to 
applications of neural memory decoding, but low-cost, comfortable EEG headsets for personal use are becoming available (e.g., the University of Oldenburg's cEEGrid \cite{debener2015}).  We plan to experiment with such devices to see 
whether we can achieve results similar to those reported here.

An especially important issue to address in application of neural memory
decoding is biometric data privacy.  Today, the risk of sharing one's EEG data 
is not well understood.  Applications using neural memory decoding can be 
designed in such a way that user EEG data never leaves a user's device.  
A potential benefit of using representation learning for neural memory
decoding is that it opens the possibility of sharing with service providers only 
the encoded representation of a user's EEG data.

\section*{Acknowledgments}
The work of Michael Haidar, and the equipment used in this project, was
partly funded by the Undergraduate Research Opportunities Center (UROC)
at California State University, Monterey Bay, the Koret Foundation, and the
Office of the Director (OD) of the National Institutes of Health under grant number R25MD010391. The content is solely the responsibility of the authors and does not necessarily represent the official views of the National Institutes of Health.

\bibliography{brain-eeg}
\bibliographystyle{ieeetr}

\newpage

\begin{appendices}

\section{Wikipedia topics}
\label{sec:appendix}

\begin{description}
\item[Set 6] Urban Planning, Alan Greenspan, Tamils, Trojan Horse, Deep Sea Exploration, Neolithic
\item[Set 7] Sea Breeze, Dublin, Scientific Method, Garage Rock
\item[Set 8] Bojack Horseman
\item[Set 9] AI Winter
\item[Set 10] Quagga, Norval Morrisseau, Moonlanding Conspiracy, Brownsville, Kabuki, Candlepin Bowling, House of Faberge, Rwanda, Pine, Nelson Mandela
\item[Set 11] Noh, Appropriation, Finland
\item[Set 12] Korean Idol, Computer Algebra, Christopher Wren, Carson City Nevada
\item[Set 13] Bronze Age, Perenial Plant, University of Calgary, Populism, William Tweed, White Blood Cells, Booker Prize, Raul Julia, New Delhi, Premier League
\item[Set 14] Algebraic Structure, Private Equity, Surf Culture, Optical Illusion, Kimono, Maglev, Monopoly, James K Polk, Copenhagen, Plant Communication
\item[Set 15] Betty Ford, Coal In China, Inorganic Chemistry, Optical Transistor, Close Up Magic, Euphrates, Tornado, Sieve of Emtostheves, Leo Tolstoy
\item[Set 16] Power Plant, US Census Bureau, Abraham Flexner, Megadiverse Countries, Environmental Migrant
\item[Set 17] Burrell Collection, Caveat Emptor, Cuban Missile Conflict, Dante Alighieri, Donald Knuth, Jacinda Arden, Nigeria, Pinball, Trans Canada Highway, Walkman Effect
\item[Set 19] Chicago Tribune, Dhaka, Everglades, International Court Of Justice, Joan Didion, Language Death, Miguel De Cervantes, Nihilism, Rhino Entertainment, Sister City
\item[Set 20] The Green Revolution, Lake Victoria, Calligraphy, Marx Brothers, Vandals, Gravity Assist, Thomas Lipton, Amaranth, Samoa, Vavilov Center
\item[Set 21] Nuremberg, Seven Years War, Atoll, Trackball, Liz Clairborne, Uluru, Hass Avocado, Folate, Tessellation, Carlo Gambino
\end{description}

\end{appendices}

\end{document}